%% file: main.tex
\documentclass[10pt,twocolumn,letterpaper]{article}

\usepackage[pagenumbers]{cvpr}              
\usepackage{xcolor}
\usepackage{color}
\usepackage{multirow}
\usepackage{listings}
\usepackage[misc]{ifsym}
\definecolor{mygreen}{RGB}{168,216,120}
\definecolor{mygray}{RGB}{218,220,225}
\definecolor{myblue}{RGB}{202,225,243}


\definecolor{cvprblue}{rgb}{0.21,0.49,0.74}
\usepackage[pagebackref,breaklinks,colorlinks,allcolors=cvprblue]{hyperref}

\definecolor{sgreen}{RGB}{30, 150, 30} 
\definecolor{sred}{RGB}{192, 0, 0}

\usepackage{makecell}
\usepackage{cutwin}

\def\paperID{92}
\def\confName{CVPR}
\def\confYear{2025}

\title{
    
\begin{cutout}{0}{3.05cm}{0pt}{1}
    {\color{white} empty} \protect\linebreak
    \protect\linebreak
    TokenFocus-VQA\ \ \ \ \ \ \ :~Enhancing Text-to-Image Alignment with Position-Aware\\ \parbox{\linewidth}{~~~~~~~~~~~~~~~~~~~~~~~ Focus and Multi-Perspective Aggregations on LVLMs}
\end{cutout}
}

\author{
Zijian Zhang$^{1*}$,~~
Xuhui Zheng$^{2*\dagger}$,~~
Xuecheng Wu$^{3\dagger}$,~~
Chong Peng$^{1\ddagger}$,~~
Xuezhi Cao$^{1}$\\
$^1$Meituan Inc;~~ 
$^2$Nanjing University;~~
$^3$Xi'an Jiaotong University\\
{\small \tt \{zhangzijian14, pengchong\}@meituan.com}
}

\begin{document}
\maketitle

\renewcommand{\thefootnote}{\fnsymbol{footnote}}
\footnotetext[1]{Equal Contribution.}
\footnotetext[2]{Work done during internship at Meituan Inc.}
\footnotetext[3]{Corresponding author.}

\input{sec/0_abstract}   
\input{sec/1intro}

\input{sec/2related_work}
\input{sec/3method}

\input{sec/4experiment}

\input{sec/5conclusion}


{
    \small
    \bibliographystyle{ieeenat_fullname}
    \bibliography{main}
}

\end{document}

%% file: sec/0_abstract.tex
\begin{abstract}
While text-to-image (T2I) generation models have achieved remarkable progress in recent years, existing evaluation methodologies for vision-language alignment still struggle with the fine-grained semantic matching. Current approaches based on global similarity metrics often overlook critical token-level correspondences between textual descriptions and visual content. To this end, we present TokenFocus-VQA, a novel evaluation framework that leverages Large Vision-Language Models (LVLMs) through visual question answering (VQA) paradigm with position-specific probability optimization. Our key innovation lies in designing a token-aware loss function that selectively focuses on probability distributions at pre-defined vocabulary positions corresponding to crucial semantic elements, enabling precise measurement of fine-grained semantical alignment. The proposed framework further integrates ensemble learning techniques to aggregate multi-perspective assessments from diverse LVLMs architectures, thereby achieving further performance enhancement. Evaluated on the NTIRE 2025 T2I Quality Assessment Challenge Track 1, our TokenFocus-VQA ranks 2nd place (0.8445, only 0.0001 lower than the 1st method) on public evaluation and 2nd place (0.8426) on the official private test set, demonstrating superiority in capturing nuanced text-image correspondences compared to conventional evaluation methods.
\end{abstract}

%% file: sec/1intro.tex
\section{Introduction}
The remarkable progress in text-to-image (T2I) generation has fundamentally transformed creative workflows, yet simultaneously exposed critical gaps in evaluation methodologies. As the generative models achieve unprecedented photorealism, the research community faces growing challenges in systematically assessing the fine-grained alignment between textual descriptions and visual content, which is a capability essential for model refinement and real-world application deployments.

Traditional evaluation paradigms have evolved from holistic quality metrics like FID \cite{heusel2017gans} and IS \cite{salimans2016improvedtechniquestraininggans} to specialized benchmarks probing specific capabilities. Evaluation frameworks such as T2I-CompBench \cite{huang2023t2i} systematically assess colors, shapes, or texture binding through the structural prompts, while REAL \cite{li2025real} evaluates visual authenticity across attributes, relationships, as well as styles. Emerging knowledge-intensive evaluations like T2I-FactualBench \cite{huang2024t2i} further verify scientific and historical accuracy, with Winoground-T2I \cite{zhu2023contrastive} examining compositional sensitivity through the contrastive examples. As the NTIRE 2025 competition, which is based on the EvalMuse-40k dataset~\cite{han2024evalmuse40kreliablefinegrainedbenchmark}, has emerged, element existence verification becomes more focused than ever before, aiming to develop specific models that can evaluate detailed image-text alignment scores more consistent and accurate with human preferences. A detailed data use case is illustrated in the \cref{fig:Data_Show} below.

\begin{figure*}
\centering
\includegraphics[width=0.782\linewidth]{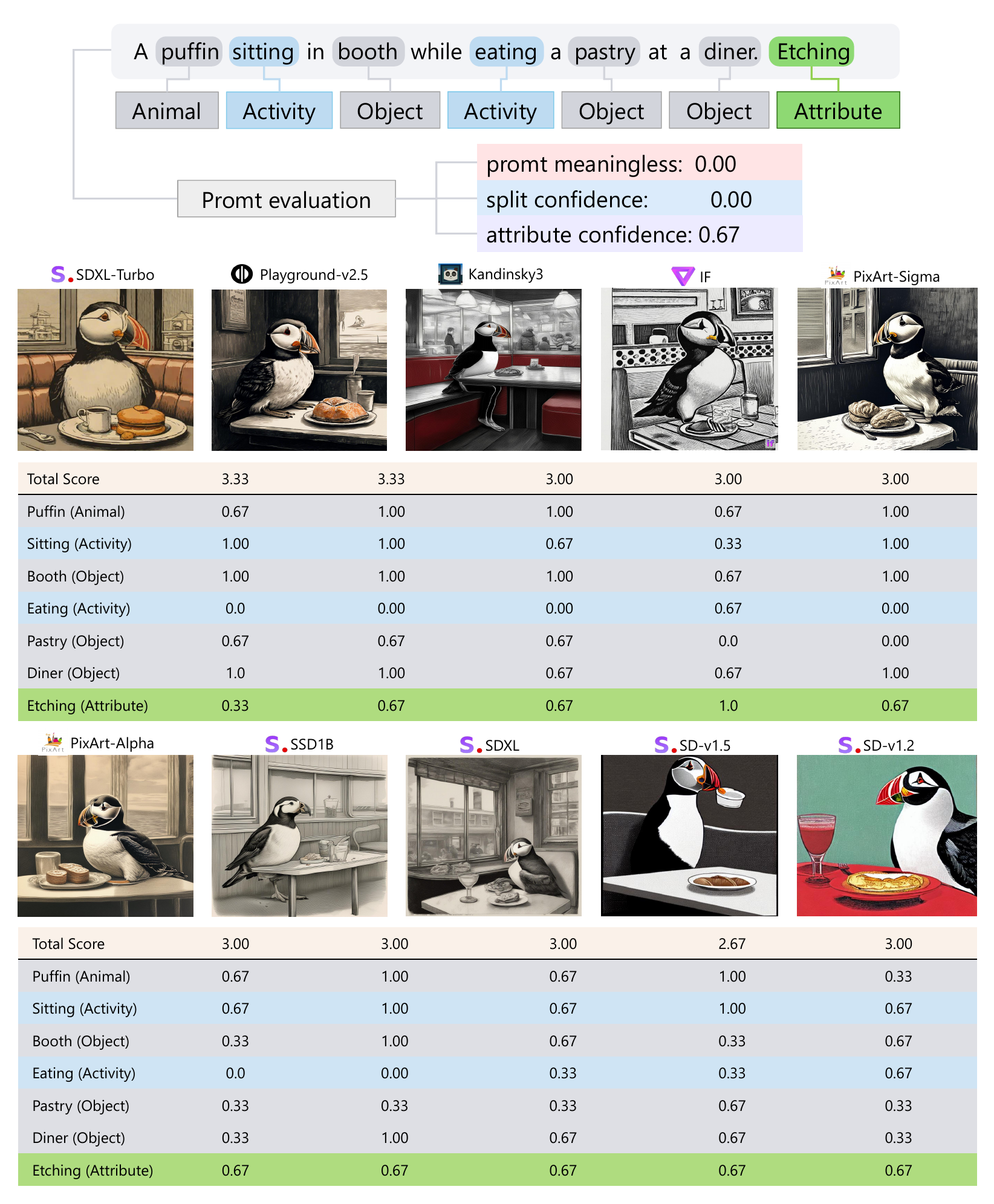}
\vspace{-0.5em}
\caption{Actual use case demonstration of the EvalMuse-40K in the  NTIRE 2025 Challenge. Different types of elements are marked with special colors (\textit{i.e.}, \textcolor{mygray}{\rule{1.2ex}{1.2ex}} for object elements, \textcolor{myblue}{\rule{1.2ex}{1.2ex}} for action elements, and \textcolor{mygreen}{\rule{1.2ex}{1.2ex}} for item attributes). The total score is classified into 1-5, and the element-level score is 0 and 1. The values shown in the tables above are the averaged results of three or six annotators.}
\vspace{-0.8em}
\label{fig:Data_Show}
\end{figure*}


Current approaches for alignment assessment primarily reveal three distinct evolutionary paths. (\textbf{I})~Global similarity metrics like CLIP Score \cite{hessel2022clipscorereferencefreeevaluationmetric} and BLIP Score \cite{li2022blip} compute image-text embedding correlations but fail to capture the token-level correspondences. (\textbf{II})~Cross-modal attention mechanisms in SCAN \cite{lee2018stacked} and ALBEF \cite{li2021align} successfully improve the localization capabilities through feature alignment, yet still struggle with the positional binding verification. (\textbf{III})~The recent paradigm shift toward VQA-based evaluation, exemplified by TIFA \cite{hu2023tifaaccurateinterpretabletexttoimage} and contemporary works \cite{li2024evaluating}, converts alignment assessment into question-answering (QA) tasks but critically overlooks probability distributions at the semantically crucial token positions. This critical limitation arises from the reliance of current methods on binary classification outputs (\textit{i.e.}, Yes or No), which discard crucial confidence information embedded in LVLMs’outputs—especially at the vocabulary positions corresponding to key objects and attributes, as extensively demonstrated in BLIP2 \cite{li2023blip} and FGA-BLIP2 \cite{han2024evalmuse40kreliablefinegrainedbenchmark}.

The above observations reveal the fundamental limitations in modern evaluation frameworks: the underutilization of position-specific probability signals, uniform treatment of all vocabulary items during the loss calculation, and inherent biases in single-model assessments. Our heuristic analysis of the VSE++ \cite{faghri2017vse} and CLIP~\cite{radford2021learningtransferablevisualmodels} architectures further demonstrates how standard similarity metrics dilute the focus on critical semantic elements during global aggregations. This technical landscape motivates TokenFocus-VQA, a framework that reshapes VQA-based evaluation through targeted probability optimization on LVLMs. Overall, the main contributions of this paper are three-fold:

\begin{itemize}
\item We first introduce the \textbf{T}oken-\textbf{F}ocus supervised and \textbf{P}osition-\textbf{S}pecific loss function to promote LVLMs fine-tuning, thereby leading to significant improvements in fine-grained image-text matching.
\item We then propose a newly optimized ensemble framework to perform multi-perspective aggregations, which integrates Bagging, Stacking and Blending, to overcome the limitations of single LVLMs in image-text evaluation.
\item Extensive experimental results demonstrate the effectiveness of our proposed TokenFocus-VQA, exhibiting impressive performance on the EvalMuse-40K dataset and the NTIRE 2025 Challenge test bed.
\end{itemize}

%% file: sec/2related_work.tex
\section{Related Works}
\subsection{Image-Text Alignment}
With the exponential growth of multimedia content on the Internet, cross-modal image-text matching has emerged as a fundamental task in information retrieval~\cite{zhang2018deep, wei2020multi}, social media analysis~\cite{zhu2022multimodal, liao2022image}, and intelligent recommendation systems~\cite{ramisa2024multi,d2024zero,nam2024dreammatcher}. Prior to the deep learning era, image feature extractions predominantly focus on the hand-crafted descriptors such as SIFT and SURF~\cite{10.1007/11744023_32, lowe2004distinctive}. While these methods have demonstrated certain effectiveness, they often suffer from limited generalization capabilities and poor adaptability to complex scenarios. The rapid advancement of deep learning has revolutionized feature extraction paradigms for both visual and textual modalities. Pioneering works like VSE++~\cite{faghri2017vse} have established baseline frameworks by optimizing cosine similarity loss between cross-modal feature representations. Subsequent methods introduce the finer-grained alignment mechanisms, exemplified by SCAN~\cite{lee2018stacked} with its stacked cross-attention modules. The introduction of dual-stream architectures have reached a milestone with ViLBERT~\cite{lu2019vilbert}, which extends the BERT~\cite{kenton2019bert} pretraining paradigm to the multimodal domain through masked multimodal data modeling.

The paradigm shift towards large-scale pre-training has yielded groundbreaking approaches like CLIP~\cite{radford2021learningtransferablevisualmodels}, which leverages the contrastive learning on 400M image-text pairs to achieve SOTA zero-shot cross-modal capabilities. Building upon visual transformer architectures, ViLT~\cite{kim2021vilt} pioneers a unified transformer framework that can directly process image patches and text tokens, enabling efficient cross-modal fusion. To balance flexibility with performance, VLMO~\cite{bao2022vlmo} proposes a mixture-of-modality-experts approach, which incorporates task-specific expert modules, enabling the model to dynamically adapt to both unimodal and multimodal tasks. To tackle data quality challenges, BLIP~\cite{li2022blip} introduces a novel architecture combining understanding and generation capabilities. The Q-Former module of BLIP-2~\cite{li2023blip} has achieved state-of-the-art visual reasoning performance through efficient cross-modal interaction learning, while maintaining computational efficiency by freezing the pretrained vision-language backbones~\cite{li2023blip}.

\subsection{Large Vision-Language Models}
In recent years, Large Vision-Language Models (LVLMs) have made significant progress in the field of multimodal understanding by integrating large-scale pretrained language models with specific vision encoders. For example, LLaVA series~\cite{liu2024visual,llavaonevision2024,liu2024improvedbaselinesvisualinstruction} models achieve precise image-text matching by directly connecting the CLIP vision encoder with the backend language model LLaMA~\cite{touvron2023llamaopenefficientfoundation} through end-to-end visual instruction tuning~\cite{liu2023visual}. InternVL~\cite{chen2024internvlscalingvisionfoundation}, by constructing a vision encoder with 6 billion parameters (ViT-6B) aligned with the language model, has achieved parameter balance between the vision and language branches for the first time, thereby overcoming the modality gap in cross-modal feature fusion~\cite{chen2024internvl}. Meanwhile, GPT-4V~\cite{2023GPT4VisionSC} and Gemini~\cite{team2023gemini}, through large-scale parameter size and multimodal instruction tuning, can support complex visual reasoning tasks~\cite{2023GPT4VisionSC, team2023gemini}. In terms of fine-grained and dynamic modeling, LLaVA-NeXT~\cite{li2024llavanextinterleavetacklingmultiimagevideo} expands capacity to a scale of 34 billion parameters, supports input with 4 $\times$ pixel resolution, and achieves general understanding across images and videos through multitask joint training. InternVL-2.5~\cite{chen2024expanding} proposes a dynamic resolution adaptation strategy, supporting multi-scale image input resolution from 224 pixel to 1024 pixel, and achieves semantic consistency across resolutions through a feature pyramid network~\cite{chen2024expanding}. Qwen-VL~\cite{bai2023qwen} introduces a textual encoding strategy for the bounding boxes, enabling spatial position awareness through extensive text labeling~\cite{bai2025qwen2}.

Compared to the traditional multimodal models, LVLMs demonstrates significant advantages in image-text alignment tasks. By employing end-to-end semantic fusion architecture and dynamic computation optimization, LVLMs has the capability to overcome the reliance of traditional models on fixed resolution input and manual feature engineering, achieving SOTA fine-grained semantic alignment across languages and scales~\cite{Ye2025ASO}. LVLMs can support multimodal autonomous reasoning, adaptive token compression, as well as zero-shot transfer learning, significantly enhancing alignment accuracy and robustness in complex scenarios such as occlusion, abstract metaphors, and multi-object interactions. In addition, through a multi-stage reasoning mechanism, LVLMs improve the efficiency of high-resolution image processing, providing more efficient solutions for practical applications such as cross-modal retrieval and multilingual matching~\cite{10.1007/978-981-96-0151-6_1, jiang2024effectiveness}.

%% file: sec/3method.tex
\begin{figure*}
\centering
\includegraphics[width=0.84\linewidth]{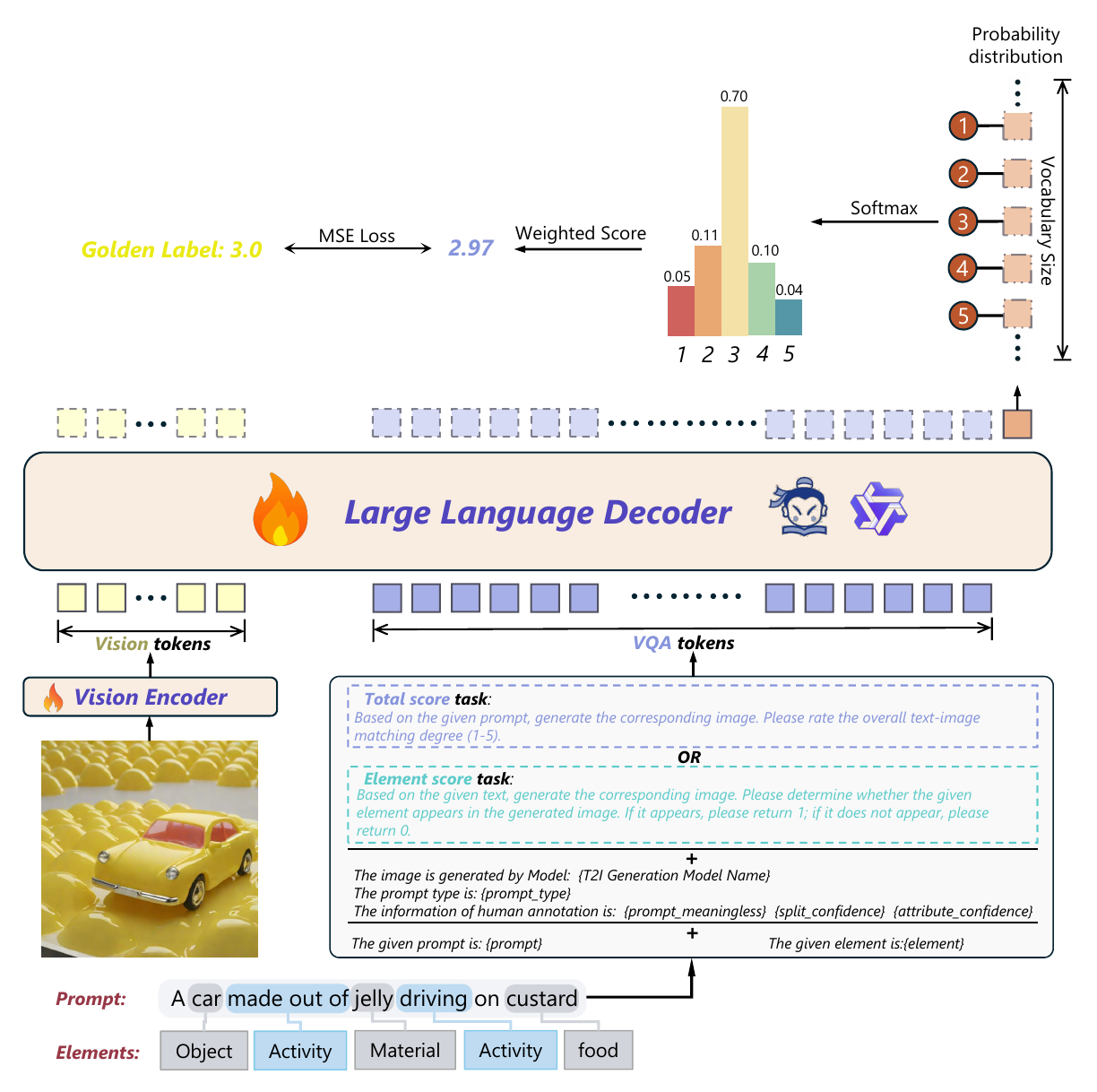}
\vspace{-0.5em}
\caption{The overall framework of our proposed TokenFocus-VQA, which is proposed for LVLMs-based T2I alignment accessment at both the holistic and fine-grained levels. The visual encoding process begins with transforming input images into the visual tokens via a vision encoder. For distinct scoring tasks (\textit{i.e.}, \textcolor[RGB]{191,195,237}{\textit{Total Score}} \& \textcolor[RGB]{128,211,211}{\textit{Element Score}}), we construct task-specific input prompts augmented with the structured meta-data. These multimodal tokens are then jointly processed in the large language decoder (\textit{i.e.}, InternLM~\cite{cai2024internlm2technicalreport} and Qwen2.5~\cite{qwen2025qwen25technicalreport}) for the generative score prediction. The framework is ultimately refined through our proposed Position-Aware Token-Focused Optimization method for further performance gains.}
\label{fig:Overall_framework}
\vspace{-0.5em}
\end{figure*}

\section{Methodology}
In this section, we introduce the T2I alignment enhanced evaluation method termed TokenFocus-VQA based on LVLMs for both holistic and fine-grained level matching. Only by deploying VQA and applying token-level supervised loss calculation during supervised fine-tuning (SFT), accurate image-text matching evaluation and recognition can be achieved at various granularities. 

As illustrated in \cref{fig:Overall_framework}, the overall framework of our proposed TokenFocus-VQA builds upon the established paradigm of VQA while introducing several critical innovations. First of all, the image and structured query are encoded into visual and textual tokens, which are then generatively understood and predicted by pre-trained LVLMs. Our key innovation emerges in the answer generation phase. Contrary to the standard VQA approaches that consider complete answer sequences, we implement \textbf{Token-Focus}, a strategic emphasis on the \textbf{first generated token} under strictly controlled output formatting. 
We then integrate the loss calculation method of numerical regression model for the label prediction dimension of language model. 
Additional multi-model integration (including serial stacking, parallel bagging, as well as hybrid blending), targeted learning rates of language and vision modalities, and other methods are also applied to further enhance performance.

\begin{table*}
\centering
\setlength{\abovecaptionskip}{0.3em} 
\setlength{\arrayrulewidth}{0.40pt}  
\renewcommand\arraystretch{1.1}      
\resizebox{\linewidth}{!}{
\begin{tabular}{l|p{14cm}}
\toprule
External Information & Detailed Descriptions \\
\midrule
T2I Model Name & The specific \textbf{model} used to perform image generation. \\

Prompt Type & The \textbf{real user prompts}, which are extensively collected from DiffusionDB~\cite{wang2022diffusiondb}, as well as the \textbf{synthetic prompts}. \\

Prompt Evaluation & The data includes \textbf{manually annotated fields}, which are deployed to evaluate both the \textbf{prompt quality}, assessing its \textbf{semantic clarity} and \textbf{generability}, and the \textbf{division clarity} of fine-grained alignment targets, including segmentation and attribute confidence. \\
\bottomrule
\end{tabular}}
\vspace{0.5em}
\caption{The detailed descriptions of the external structural information.}
\label{tab:External structural information}
\vspace{-1em}
\end{table*}

\subsection{Position-Aware Token-Focused Optimization}
Language models (LMs) generate probabilistic outputs via softmax, which inherently conflicts with deterministic regression tasks (\textit{i.e.}, numerical scoring) requiring focus on specific tokens. Standard cross-entropy supervision forces probability mass allocation across all tokens, diluting learning signals and slowing convergence. To address this, we propose \textbf{token-focus} supervision, re-weighting the loss to concentrate on task-critical tokens. This filters extraneous noise and transforms probabilistic training into value-driven optimization, directly aligning LM generation with continuous regression metrics.
Specifically, we only focus on the first generated token and obtain the \textbf{position-aware} probability of the label corresponding to the score ([0, 1] or [1, 2, 3, 4, 5]) in its predicted distributions. After normalization, we then multiply the probability of the corresponding label by the score weight to obtain the LVLMs-based regression or classification results, and deploy MSE (Mean-Square Error) and other methods to calculate the loss accordingly. 

Let the language model vocabulary be $V$, the target score set be $S=\{s_1, s_2, \dots, s_k\}$ ($[1, 2, \dots, 5]$ for element score tasks, $[1, 2]$ for total score task, $k$ for score label nums). Given an input image-text pair $X$, the model's original probability distribution for the first generated (t = 1) token can be formulated as:
\begin{equation}
    p_{t=1}(w|X) = \text{softmax}(z_w), \quad \forall w \in V ,
\end{equation}
where $z_w$ represents the output value of token $w$ from the last output linear layer. After filtering irrelevant tokens, we can get the conditional probability distribution after normalization (\textit{i.e.}, Softmax function):
\begin{equation}
    P(s_i) = \frac{exp(p_{t=1}(s_i|X))}{\sum_{j=1}^k exp(p_{t=1}(s_j|X))}, \quad s_i \in S. 
\end{equation}

This operation projects the original probability space into the target score space to eliminate potential noise interference. Then the discrete-to-continuous conversion of predicted value $\hat{y}$ is achieved through the expected value mapping, \textit{i.e.},
\begin{equation}
    \hat{y} = \mathbb{E}_{s \sim P}[s] = \sum_{i=1}^k s_i P(s_i). 
\end{equation}

The MSE is deployed to directly optimize the gaps between the predicted value and true value, the loss of any task ($\mathcal{L}_{\text{task}} $) can be calculated as:
\begin{equation}
    \mathcal{L}_{\text{task}} = (\hat{y}_n - y_n)^2,
\end{equation}
where $\hat{y}_n$ and $y_n$ refers to the final predictions and ground-truth, respectively.

\subsection{External Structural Information Integration}
\label{sec:External structural information integration of image-text pairs}

Considering that prompt engineering can effectively enhance the performance of Large Language Models (LLMs) on specific tasks, and inspired by the common practice of leveraging additional features to improve recognition accuracy in machine learning, we propose a structured prompt construction method specifically designed for image-text pairs in VQA tasks. It systematically incorporates textualized external information as engineered prompt features, as illustrated in \cref{tab:External structural information}. Our feature augmentation is motivated by two considerations, which are detailed as below:
\begin{itemize}
    \item Given the potentially significant disparities in generative capabilities across different model architectures, we further integrate detailed model-specific information into the prompts to compensate for the model discrepancies.
    \item The quality of the generated prompt itself will directly affect the effects of subsequent generation and interfere with the model's understanding of complex or ambiguous language. The spatial description of fine-grained elements that need to be judged will also affect the modeling ability of complex scenes. The accuracy of attributes directly guides the upper limit of model detail evaluation.
\end{itemize}

\begin{figure}[t]
\centering
\includegraphics[width=1.0\linewidth]{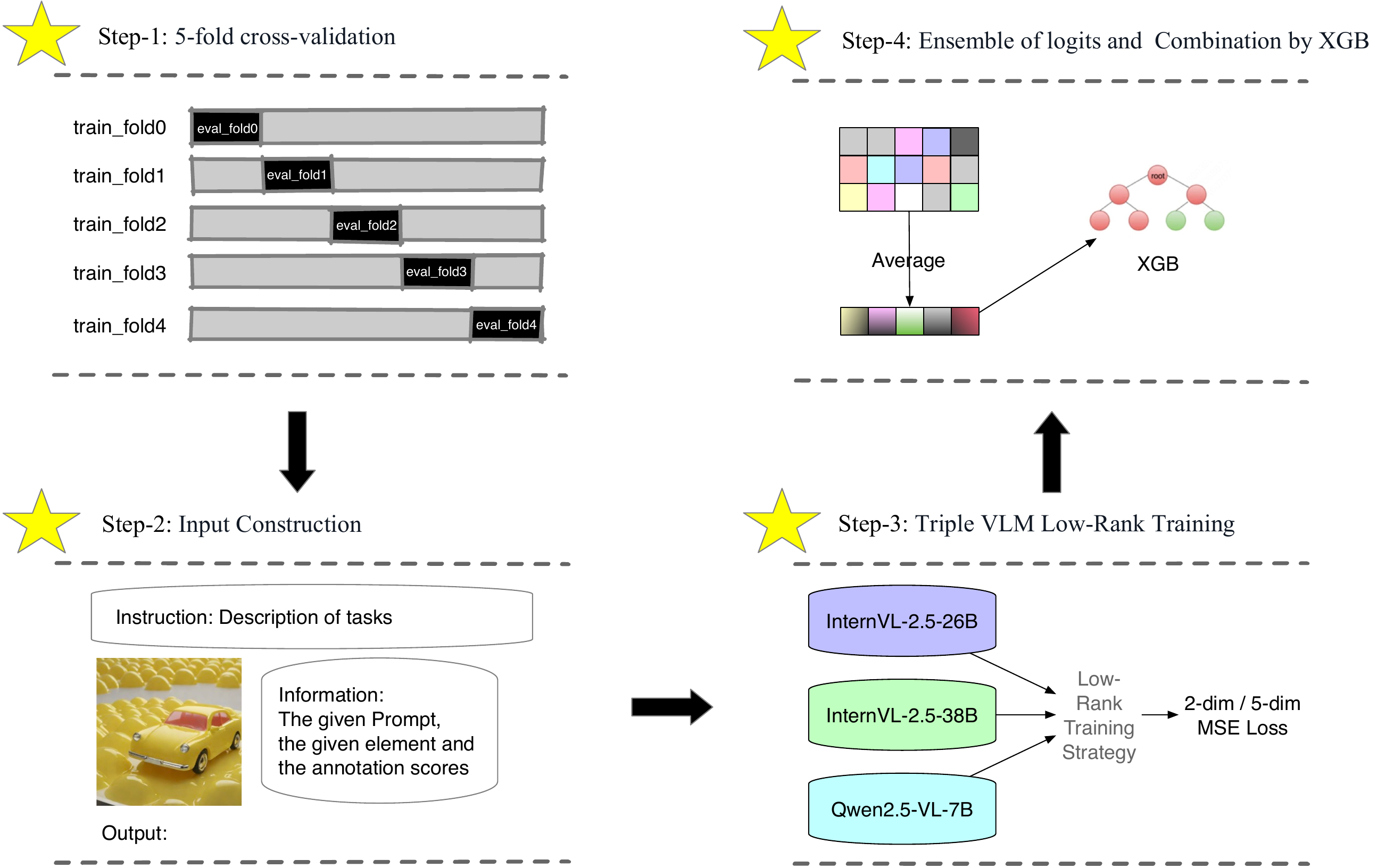}
\caption{The overall illustration of our ensemble training and inference workflow.}
\label{fig:workflow}
\vspace{-1.1em}
\end{figure}

\subsection{Data Sub-packaging and Model Ensemble}
\label{sec:Data subpackaging and model ensemble}
As displayed in \cref{fig:workflow}, we introduce a novel hierarchical ensemble architecture which systematically integrates ensemble skills to establish multi-strategy consensus formation, effectively addressing heterogeneous representation learning bottlenecks and distributional bias inherent in singular LVLM for cross-modal evaluation tasks.
Our training procedure consists of three keypoints, \textit{i.e.}, (\textbf{I}) We partition the dataset into five folds, allocating 80\% for training and 20\% for validation within each fold. Our guiding principle is to eliminate duplicate prompts, while allowing image generation models to overlap across folds, aligning with the testing data distribution. Then in each fold, an individual model is independently trained and subsequently employed for ensemble blending purposes. The biased predictions from each constituent model are aggregated into a meta-learner to enhance overall predictive efficacy through the ensemble refinement. (\textbf{II}) We modify the configuration of the model input, utilizing the prompt construction method mentioned in \cref{sec:External structural information integration of image-text pairs} to integrate certain statistical features directly as the textual inputs. 
(\textbf{III}) For the testing procedure, we utilize various models to predict the test data by deploying checkpoints derived from the training phase across different folds. In the end, we employ XGBoost~\cite{chen2016xgboost} to integrate the predicted scores from models of varying scales and architectures with selected statistical features, jointly consolidating them into the final predictions.

The annotations in the dataset represent averages from multiple annotators, and utilizing them directly without bucketing has shown superior results for the overall score task. For the element score task, the variation between employing MSE and cross-entropy is minimal. 
Our empirical analysis reveals that while full-parameter fine-tuning achieves modest gains (+0.5 pp) in localized 5-fold validation, but exhibits critical generalization deficits (-0.012 SRCC) on the leaderboard, suggesting inherent limitations in data-constrained scenarios. This observation motivates our adoption of the LoRA~\cite{hu2022lora} adaptation.

\begin{table}[t]
\centering
\setlength{\abovecaptionskip}{0.3em}  
\setlength{\arrayrulewidth}{0.40pt}   
\renewcommand\arraystretch{0.95}      
\resizebox{0.8\linewidth}{!}{
\begin{tabular}{l|c}
\toprule
Configurations & Value \\
\midrule
Optimizer & AdamW \\
LoRA rank & 64\\
LoRA alpha & 128\\
Base learning rate & 1\textit{e}-4 \\
Vision learning rate & 1\textit{e}-5 \\
Weight decay & 0.05 \\
Optimizer momentum & $\beta_1, \beta_2 = 0.9, 0.95$ \\
Global batch size & 64 \\
Learning rate schedule & cosine decay \\
Warmup steps & 200 \\
Seed & 1234\\
Epoch & 3 \\
\bottomrule
\end{tabular}}
\vspace{0.5em}
\caption{The detailed model training settings of our introduced TokenFocus-VQA. To faithfully ensure consistency, the same training setup are deployed for our LVLMs of varying architectures and scales.}
\label{tab:parameter detail}
\vspace{-1.5em}
\end{table}

%% file: sec/4experiment.tex
\section{Experiments}
\subsection{Implementation Details}
We split the overall data into 5 non-overlapping folds, selecting four folds for training each time and the rest for cross-validation, making sure there is no data overlap between prompts when splitting. We deploy Qwen2.5-VL~\cite{bai2025qwen2} and InternVL-2.5~\cite{chen2024expanding} as the baseline models to perform extensive model training.

\begin{table}[t]
\centering
\setlength{\abovecaptionskip}{0.3em}  
\setlength{\arrayrulewidth}{0.40pt}   
\renewcommand\arraystretch{1.00}      
\resizebox{1.0\linewidth}{!}{
\begin{tabular}{l|cccc}
\toprule
Method & \makecell{Visual \\ Enc.} & \makecell{Text \\ Enc.} & PLCC~($\uparrow$)  & SRCC~($\uparrow$)  \\
\midrule 
Qwen2.5-VL-7B~\cite{bai2025qwen2} (VQA) & 660M & 7B & 0.6796 & 0.6783 \\
InternVL-2.5-4B~\cite{chen2024expanding} (VQA) & 300M & 4B & 0.6922 & 0.7054 \\
CLIP-Score~\cite{hessel2022clipscorereferencefreeevaluationmetric} & 88M & 63M & 0.3023 & 0.2975 \\
BLIPv2Score~\cite{hessel2022clipscorereferencefreeevaluationmetric} & 300M & 2.7B & 0.3621 & 0.3381 \\
FGA-BLIP2\cite{han2024evalmuse40kreliablefinegrainedbenchmark} & 300M & 2.7B & 0.7754 & 0.7741 \\

\midrule
Qwen2.5-VL-7B~\cite{bai2025qwen2} & 660M & 7B & 0.7962 & 0.8020 \\
Qwen2.5-VL-32B~\cite{bai2025qwen2} & 660M & 32B & 0.7977 & 0.8007 \\

\midrule
InternVL-2.5-4B~\cite{chen2024expanding} & 300M & 3B & 0.7988 & 0.8025 \\
InternVL-2.5-8B~\cite{chen2024expanding} & 300M & 7B & 0.8003 & 0.8046 \\
InternVL-2.5-26B~\cite{chen2024expanding} & 6B & 20B & 0.8096 & \textbf{0.8141} \\
InternVL-2.5-38B~\cite{chen2024expanding} & 6B & 32B & \textbf{0.8098} & 0.8133 \\

\bottomrule
\end{tabular}}
\vspace{0.5em}
\caption{The experimental results of different LVLMs using one fold data without extra structural information. VQA stands for applying VQA method on LVLMs and Enc. refers to Encoder, which is deployed to compare the size of Vision encoder and Text encoder. InternVL-2.5~\cite{chen2024expanding} series features multiple vision encoder variants (\textit{i.e.}, 300M and 6B parameters) for scalable deployment, while the Qwen2.5-VL~\cite{bai2025qwen2} series maintains architectural uniformity with a fixed 660M visual encoder across all configurations.}
\label{tab:The experimental results of different model using one fold data}
\vspace{-0.4em}
\end{table}

\begin{table}[t!]
\centering
\setlength{\abovecaptionskip}{0.3em}  
\setlength{\arrayrulewidth}{0.40pt}   
\renewcommand\arraystretch{1.00}      
\resizebox{0.8\linewidth}{!}{
\begin{tabular}{l|ccc}
\toprule[1pt]     
Fold & SRCC~($\uparrow$) & PLCC~($\uparrow$) & ACC~($\uparrow$) \\
\midrule   
=~1 & \textbf{0.8371} & \textbf{0.8313}  & 82.35\% \\
=~2 & 0.8213 & 0.8184  & 82.06\% \\
=~3 & 0.8175 & 0.8144  & \textbf{82.40\%} \\
=~4 & 0.8272 & 0.8226  & 82.32\% \\
=~5 & 0.8163 & 0.8122  & 81.72\% \\
\bottomrule[1pt]     
\end{tabular}}
\vspace{0.5em}
\caption{The performance comparisons of 5-fold cross validation utilizing the InternVL-2.5-26B~\cite{chen2024expanding}.}
\label{tab:fold_exp}
\vspace{-1.3em}
\end{table}

\noindent \textbf{Training Settings: }Different learning rates are set for the vision encoder and LM decoder layer to balance the emphasis on visual understanding and task instruction following. The LoRA~\cite{hu2022lora} is applied for efficient fine-tuning while conducting extensive training experiments on different models of varying sizes. The specific training parameters are shown in \cref{tab:parameter detail}. All the experiments are conducted on the \textit{NTIRE 2025 Text to Image Generation Model Quality Assessment Challenge Track 1} dataset using a machine with 8 $\times$ NVIDIA A100 GPUs, with respective training durations of 7 hours (Total Scoring) and 25 hours (Element Scoring) under a standardized 5-fold cross-validation protocol with independent optimization across splits.

\begin{table}[t!]
\centering
\setlength{\abovecaptionskip}{0.3em}  
\setlength{\arrayrulewidth}{0.40pt}   
\renewcommand\arraystretch{1.00}      
\resizebox{0.8\linewidth}{!}{
\begin{tabular}{l|ccc}
\toprule[1pt]     
Fold & SRCC~($\uparrow$) & PLCC~($\uparrow$) & ACC~($\uparrow$) \\
\midrule   
=~1 & 0.8273 & 0.8260 & 81.78\% \\
=~2 & 0.8189 & 0.8190 & 81.63\% \\
=~3 & 0.8163 & 0.8155 & 81.49\% \\
=~4 & 0.8233 & 0.8218 & 81.80\% \\
=~5 & 0.8191 & 0.8172 & 81.61\% \\
\midrule
Avg & 0.8210 & 0.8199  & 81.66\% \\
Blend & 0.8317 & 0.8302  & 82.36\% \\
\bottomrule[1pt]     
\end{tabular}}
\vspace{0.5em}
\caption{The performance comparisons of the ensemble strategy on public leaderboard using InternVL-2.5-26B. Blend refers to an ensemble strategy where predictions from all cross-validation folds are aggregated as input features for a tree-based model, without leveraging supplementary structural metadata.}
\label{tab:fold_ensemble}
\vspace{-1.3em}
\end{table}

\begin{table*}[t!]
\centering
\setlength{\abovecaptionskip}{0.3em}  
\setlength{\arrayrulewidth}{0.40pt}   
\renewcommand\arraystretch{1.00}      
\resizebox{\textwidth}{!}{
\begin{tabular}{p{2mm}p{35mm}p{30mm}p{25mm}p{25mm}p{25mm}p{25mm}}
\toprule     
ID & ~~~~~~~Method & \hspace{0.05em}Evaluation~Bed  & SRCC & PLCC & ~ACC & \hspace{-0.3em}Overvall \\
\midrule

\multirow{2}{=}{0} & \multirow{2}{=}{Qwen2.5-VL-7B} & \multirow{1}{=}{Cross Validation} & \multirow{1}{=}{0.8256} & \multirow{1}{=}{0.8205} & \multirow{1}{=}{0.8252} & \multirow{1}{=}{0.8241} \\
        & & \multirow{1}{=}{~~Leaderboard} & \multirow{1}{=}{~~~~\textbf{--}} & \multirow{1}{=}{~~~~\textbf{--}} & \multirow{1}{=}{~~~~\textbf{--}} & \multirow{1}{=}{~~~~\textbf{--}} \\
\midrule

\multirow{2}{=}{1} & \multirow{2}{=}{InternVL-2.5-26B} & \multirow{1}{=}{Cross Validation} & \multirow{1}{=}{0.8258} & \multirow{1}{=}{0.8198} & \multirow{1}{=}{0.8217} & \multirow{1}{=}{0.8223}\\
         & & \multirow{1}{=}{~~Leaderboard} & \multirow{1}{=}{\textcolor{sgreen}{0.7839}} & \multirow{1}{=}{\textcolor{sgreen}{0.8125}} & \multirow{1}{=}{\textcolor{sgreen}{0.8509}} & \multirow{1}{=}{\textcolor{sgreen}{0.8245}}\\
        \midrule
        
        \multirow{2}{=}{2} & \multirow{2}{=}{InternVL-2.5-38B} & \multirow{1}{=}{Cross Validation} & \multirow{1}{=}{0.8273} & \multirow{1}{=}{0.8232} & \multirow{1}{=}{0.8226} & \multirow{1}{=}{0.8239}\\
        & & \multirow{1}{=}{~~Leaderboard} & \multirow{1}{=}{~~~~\textbf{--}} & \multirow{1}{=}{~~~~\textbf{--}} & \multirow{1}{=}{~~~~\textbf{--}} & \multirow{1}{=}{~~~~\textbf{--}}\\
        \midrule

        \multirow{2}{=}{3} & \multirow{2}{=}{Qwen2.5-VL-7B \\ + InternVL-2.5}
        & \multirow{1}{=}{Cross Validation} & \multirow{1}{=}{~~~~\textbf{--}} & \multirow{1}{=}{~~~~\textbf{--}} & \multirow{1}{=}{~~~~\textbf{--}} & \multirow{1}{=}{~~~~\textbf{--}}\\
        & & \multirow{1}{=}{~~Leaderboard} & \multirow{1}{=}{0.8002 (\textcolor{sred}{+0.0163})} & \multirow{1}{=}{0.8321 (\textcolor{sred}{+0.0196})} & \multirow{1}{=}{0.8619 (\textcolor{sred}{+0.0110})} & \multirow{1}{=}{0.8390 (\textcolor{sred}{+0.0145})}\\
        
        \midrule

        \multirow{2}{=}{4} & \multirow{2}{=}{Qwen2.5-VL-7B + InternVL-2.5 $\spadesuit$} & \multirow{1}{=}{Cross Validation} & \multirow{1}{=}{~~~~\textbf{--}} & \multirow{1}{=}{~~~~\textbf{--}} & \multirow{1}{=}{~~~~\textbf{--}} & \multirow{1}{=}{~~~~\textbf{--}}\\
        & & \multirow{1}{=}{~~Leaderboard} & \multirow{1}{=}{0.8002 (\textcolor{sred}{+0.0163})} & \multirow{1}{=}{0.8321 (\textcolor{sred}{+0.0196})} & \multirow{1}{=}{0.8691 (\textcolor{sred}{+0.0182})} & \multirow{1}{=}{0.8426 (\textcolor{sred}{+0.0181})}\\
        \bottomrule
\end{tabular}}
\vspace{0.5em}
\caption{The performance comparisons of different methods in terms of SRCC, PLCC, ACC, as well as Overvall metrics on both 5-fold Cross Validation and Leaderboard evaluation beds. InternVL-2.5: InternVL-2.5-26B \& InternVL-2.5-38B, $\spadesuit$: Statistic Features. \textcolor{sgreen}{\textit{Green}} refers to the baseline results for longitudinal comparison, representing the non-ensemble learning-enhanced approach. \textcolor{sred}{\textit{Red}} denotes the performance enhancement.}
\label{tab:merged_rows}
\vspace{-0.8em}
\end{table*}

\begin{table}[t!]
\centering
\setlength{\abovecaptionskip}{0.3em}  
\setlength{\arrayrulewidth}{0.40pt}   
\renewcommand\arraystretch{1.00}      
\resizebox{0.6\linewidth}{!}{
\begin{tabular}{l|cc}
\toprule    
Fold  &T-Samples &  E-Samples\\
\midrule   
=~1 &26,191 & 6,526\\
=~2 &26,099 & 6,618\\
=~3 &26,164 & 6,553\\
=~4 &26,184 & 6,533\\
=~5 &26,245 & 6,472\\
\bottomrule     
\end{tabular}}
\vspace{0.5em}
\caption{The detailed informaction on data fold splitting. T and E refer to Training and Evaluation, respectively. The overall data is devided according to the unique prompt ID, maintaining a 4~:~1 ratio (2,393~:~598) of unique prompts between training and evaluation sets in each fold.}
\label{tab:fold_data_details}
\vspace{-1.2em}
\end{table}

\noindent \textbf{Evaluation Settings: } For the overall alignment scores, we report the Spearman Rank Correlation Coefficient (SRCC) and Pearson Linear Correlation Coefficient (PLCC) to measure the correlation between model predictions and human annotations. We further conduct fine-grained elements evaluation by reporting the accuracy (ACC) of the output predictions. To further ensure equitable weighting of both holistic alignment measurements and granular element matching in the comprehensive evaluation framework, we formulate the composite evaluation metric through the following mathematically formalized weighted integration:
\begin{equation}
O = 0.25 \times S + 0.25 \times P + 0.5 \times A,
\end{equation}
where $S$, $P$, and $A$ represent SRCC, PLCC, and ACC, respectively. $O$ denotes the final composite evaluation metric.

\subsection{Overall Performance Comparisons}
\label{sec:Performance of different LVLMs}
We establish comprehensive comparative baselines utilizing FGA-BLIP2~\cite{han2024evalmuse40kreliablefinegrainedbenchmark}, CLIP-Score~\cite{hessel2022clipscorereferencefreeevaluationmetric}, and BLIPv2Score~\cite{hessel2022clipscorereferencefreeevaluationmetric}. As for our experimental protocol initiates with preliminary validation on a held-out validation fold to benchmark performance variations across model architectures and scales. We evaluate two top-performing open-source models (\textit{i.e.}, Qwen2.5-VL~\cite{bai2025qwen2} and InternVL-2.5~\cite{chen2024expanding}) across parameter scales ranging from 4B to 38B. Besides, we have conducted controlled experiments employing VQA-typical implementations on LVLMs, to verify the methodological superiority of our proposed approach.

As shown in \cref{tab:The experimental results of different model using one fold data} above, our TokenFocus-VQA demonstrates statistically significant superiority over both conventional VQA approaches and the SOTA FGA-BLIP2~\cite{hessel2022clipscorereferencefreeevaluationmetric}. InternVL-2.5 consistently outperforms its counterpart in cross-modal alignment accuracy at comparable parameter scales. Remarkably, increasing LVLMs parameters through decoder expansion demonstrates negligible performance impact (\textit{e.g.}, Qwen2.5-VL-32B shows minimal PLCC improvement with SRCC degradation, a pattern replicated in InternVL-2.5 variants). This phenomenon significantly underscores the decisive role of vision encoder capacity -- scaling InternVL-2.5's vision encoder from 300M to 6B parameters yields $\approx 1\%$ absolute performance improvement, revealing the vision-centric scaling laws in multimodal systems on the image-text alignment evaluation task.


\subsection{Performance of Different Folds}


We present a comprehensive documentation of our experimental protocol through \cref{tab:fold_data_details}, which specifies the implementation details of our 5-fold cross-validation strategy employing a stratified partitioning mechanism based on unique prompt identifiers. Each prompt ID in this configuration corresponds to multiple annotated samples systematically generated by diverse text-to-image generation models, ensuring balanced representation of heterogeneous data distributions across validation folds. Following the empirical evidence from preliminary investigations demonstrating the superior baseline performance of scaled vision-language architectures (particularly InternVL-2.5~\cite{chen2024expanding} with its expanded encoder capacity), we establish this architecture as our foundational reference model for subsequent comparative analyses. The quantitative outcomes detailed in \cref{tab:fold_exp} systematically demonstrate the performance enhancements achieved through our proposed \textbf{E}xternal \textbf{S}tructural \textbf{I}nformation \textbf{I}ntegration \textbf{P}rompting \textbf{S}trategy (introduced in \cref{sec:External structural information integration of image-text pairs}) across both comprehensive alignment metrics and component-level evaluation tasks. Our proposed method yields statistically robust improvements over conventional baseline approaches examined in \cref{sec:Performance of different LVLMs}, evidenced by significant gains in holistic correlation measures and granular component recognition accuracy. These advancements effectively address the inherent limitations of standard visual question answering (VQA) paradigms that typically suffer from \textbf{insufficient} contextual grounding and \textbf{incomplete} semantic representation in the cross-modal alignment tasks, thereby establishing our framework as a robust evaluation paradigm for multimodal systems.

However, the substantial performance variance across 5-fold validations suggests that limited prompt diversity and inadequate sample cardinality can significantly induce non-negligible inter-partition discrepancies when conducting stratified data splitting.

\subsection{Performance of Ensemble Stragety}

To systematically investigate the discriminative capabilities of cross-validated model variants while preserving their complementary strengths in multimodal comprehensive understanding, we operationalize the ensemble protocol described in \cref{sec:Data subpackaging and model ensemble}, achieving statistically significant enhancement through differential weighting of vision-language attention patterns across different validation folds.

The comprehensive experimental results presented in \cref{tab:merged_rows} demonstrate the methodological progression of our hierarchical integration framework, which operates through three coordinated phases: First of all, the strategic partitioning of heterogeneous large vision-language models (LVLMs) including Qwen2.5-VL-7B~\cite{bai2025qwen2} and scaled InternVL-2.5 variants (26B \& 38B)~\cite{chen2024expanding} via 5-fold cross-validation; Second, the implementation of stacked generalization through gradient-boosted tree models that optimally combine base learners' predictions; Third, the refinement through structural-information augmented prompting. This tripartite architecture achieves metric improvements of SRCC (+0.163), PLCC (+0.196) in comprehensive alignment evaluation, coupled with +1.10\% accuracy gain in fine-grained element analysis - collectively establishing impressive overall performance improvement.  The performance trajectory further ascends through our proposed integration of external structural features into prompt engineering, which introduces additional statistically consistent enhancements across all evaluation axes by better modeling the latent relationships between semantic hierarchies and visual compositions. This validation confirms the superiority of our three-pillar ensemble philosophy: 1) Architectural diversification through complementary base model selection, 2) Meta-knowledge distillation via multi-layer stacked generalization, and 3) Cross-modal refinement through structurally-informed prompt optimization.

Overall, these innovations culminate in a highly effective framework that not only addresses the limitations of existing methods but also sets a new benchmark for cross-modal alignment evaluation tasks. Our approach demonstrates the potential of ensemble learning and structural integration to push the boundaries of model performance in fine-grained vision-language matching.

%% file: sec/5conclusion.tex
\section{Conclusion and Prospect}
This work aims to tackle the critical challenge of fine-grained vision-language alignment evaluation in text-to-image generation. By introducing TokenFocus-VQA, we establish a new evaluation framework that combines token-aware probability optimization with multi-model ensemble strategies. The proposed position-specific loss calculation enables precise supervision for localized semantic matching, while the systematic integration of Bagging, Stacking, as well as Blending techniques further enhances the evaluation robustness. The extensive experimental results on the \textit{NTIRE 2025 Text to Image Generation Model Quality Assessment Challenge} demonstrates state-of-the-art performance on public evaluations. Our proposed framework not only advances the methodological foundation for T2I (Text-to-Image) quality assessment but also provides actionable insights for advancing the semantically-aware evaluation systems in multimodal AI research.

In the future developments, we plan to explore the dynamic vocabulary adaptation and more advanced cross-modal interaction components for broader  applicability.